\documentclass{article}

\usepackage[preprint]{neurips_2019}

\usepackage[utf8]{inputenc}
\usepackage[T1]{fontenc}
\usepackage{hyperref}
\usepackage{url}
\usepackage{booktabs}
\usepackage{amsfonts}
\usepackage{nicefrac}
\usepackage{microtype}
\usepackage{graphicx}
\usepackage{xcolor}
\usepackage{lipsum}
\usepackage{amsmath}

\title{
  Neologism Learning as a Parameter-Efficient Alternative to Fine-Tuning for Model Steering \\
}

\author{
  Sungjoon Park \\
  Department of Computer Science \\
  Columbia University \\
  \texttt{sp4050@columbia.edu} \\
  \And
  Varun Ramamurthi \\
  Department of Computer Science \\
  Columbia University \\
  \texttt{vzr2104@columbia.edu} \\ 
  \And
  Owen Terry \\
  Department of Computer Science \\
  Columbia University \\
  \texttt{okt2002@columbia.edu} \\
}

\begin{document}

\maketitle

\begin{abstract}
In language modeling, neologisms are new tokens trained to represent a concept not already included in a given model’s vocabulary. Neologisms can be used to encourage specific behaviors in models, for example by appending prompts with "Give me a \{neologism\} answer." Behavioral steering can also be achieved through fine-tuning, albeit with more compute and less flexibility: learning a neologism only trains $d$ parameters and allows the user to still access the model's default behavior. We compare the performance of neologism learning against low-rank adaptation (LoRA) fine-tuning, finding that neologisms outperform fine-tuned models under a matched training setup (same data and hyperparameters). We also investigate self-verbalizations of neologisms, and observe that the model will occasionally make up its own new words when asked about a neologism.
\end{abstract}

\section{Introduction}

In natural language, new words, or neologisms, are often invented to describe new, useful concepts. Motivated by this fact and by recent work in training neologisms on language models (Hewitt et al., 2025b)\cite{hewitt25b}, we study the efficacy of using neologisms to steer the behavior of language models.

We begin by training a pair of neologisms on Mistral 7B Instruct, following a method put forward by Hewitt et al. (2025b)\cite{hewitt25b}. In neologism learning, we add a new token to the model's vocabulary, freeze all parameters except for the new token's embedding, then train on data that demonstrates the intended meaning of the neologism. Our first neologism, "\textasciitilde short", is trained to optimize for responses that are less than 50 words; our second, "\textasciitilde kidmode", is trained on synthetic responses geared towards grade-schoolers.

We test our learned neologisms by comparing the model's responses to prompts appended with "Give me a \{neologism\} answer" to responses in the training data, as well as to base model responses. For "\textasciitilde short", we evaluate based on answer length; for "\textasciitilde kidmode", we use LLM-as-a-judge to measure the extent to which each answer aligns with the desired behavior.

Next, we fine-tune the same base model on the same response data that we trained the neologisms on. Fine-tuning is more costly than neologism training. The model only learns $d=4,096$ new parameters (in this case) to train a single new token, while fine-tuning learns far more: $425,984$ parameters per LoRA rank under Mistral 7B Instruct. Furthermore, the fine-tuned model is in some sense less capable than the model that only learned the neologism, as it can only respond in the manner X it was trained on. The model with the neologism can either be prompted to respond in manner X or simply used normally. Because of this, if a desired behavior can be encoded in a neologism with sufficient accuracy, it may be preferable in certain applications to do so, rather than encoding this behavior through fine-tuning.

We fine-tune one model that gives "\textasciitilde short" responses and one model that give "\textasciitilde kidmode" responses, using rank-8 LoRA for each. We use the same training setup to fine-tune as we do to train neologisms, i.e. the same data and hyperparameters. We evaluate the fine-tuned models the same way we test the learned neologisms. Comparing, we find that under our matched training setup, neologisms outperform fine-tuning.

We also experiment with self-verbalization, a concept introduced in Hewitt et al. (2025b)\cite{hewitt25b} wherein the user asks the model questions like "What does \{neologism\} mean?" We find that the model occasionally invents previously unseen word compositions such as "mutexpoitary" and "poornessily" to describe higher-level concepts. 

\section{Related Work}

\subsection{Neologism Learning}

Neologism learning was introduced by Hewitt et al. (2025a)\cite{hewitt25a}. This position paper argues that interpretability is a communication problem, in that there are human-native concepts that machines don't easily understand, and vice versa. Through neologisms, we can build a vocabulary to bridge this gap, giving ourselves ways to communicate subtle human concepts to AIs and allowing AIs to more clearly present ideas that don't fit neatly into natural human language. 

Hewitt et al. (2025b)\cite{hewitt25b} builds on this by demonstrating how neologisms can help shed light on machine concepts through self-verbalization, i.e. asking questions like "What does \{neologism\} mean?" Interestingly, this direction of inquiry can yield "machine-only synonyms", which are internally consistent for the model but make little sense to us. For example, a model was given a neologism trained to generate single-sentence answers, and claimed that "lack" was a synonym of this neologism. Indeed, appending "Give me a lack answer" to questions caused substantially shorter responses.\cite{hewitt25b}

Zheng et al. (2024)\cite{zheng24} studies how human neologisms degrade LLM performance. New words are invented naturally over time, contributing to temporal data drift. The paper finds that machine translation performance is nearly halved on average when a single word is replaced by a nearly-synonymous neologism.

\subsection{Out-of-context reasoning}

Recent work has shown that LLMs are capable of connecting distinct concepts in surprising ways. Cloud et al. (2025)\cite{cloud25} finds that models can reliably transmit information in seemingly meaningless sequences of numbers: a model told to generate numbers imbued with a love for owls creates data that can train another model to love owls. Betley et al. (2025a)\cite{betley25a} finds that models fine-tuned implicitly on certain behaviors can explicitly articulate these behaviors. Betley et al. (2025b)\cite{betley25b} and Taylor et al. (2025)\cite{taylor25} find that fine-tuning on narrowly misaligned concepts such as insecure code and reward-hacking cause models to become broadly misaligned. Betley et al. (2025c)\cite{betley25c} finds that training a model on Hitler's benevolent traits, even without mentioning him, causes the model to adopt a Hitler persona. Collectively, these results are evidence of a gap between human and machine concepts, one that is possibly growing as LLMs become more capable.

\subsection{Steering}

Fine-tuning methods, including low-rank adaptation (LoRA) (Hu et al., 2021)\cite{hu21}, encourage models to exhibit specific desired behaviors. Other model steering tools include sparse autoencoders (SAEs) (Cunningham et al., 2023)\cite{cunningham23} and steering vectors (Tan et al., 2024)\cite{tan24}. Neologisms may provide a more flexible way to steer models; motivated by this possibility, we compare neologism steering performance against LoRA in this paper.

\section{Approach}

\subsection{Neologism learning}
\label{ref:4.1}

We train neologisms following \cite{hewitt25b}. Neologism learning consists of adding a new token to the vocabulary, freezing all parameters except for those of the new token, and training on data that exemplifies the meaning of the new token.

Specifically, we expand our model's vocabulary by one element and its embedding matrix by one column, getting $V' = V \cup \{c\}, E \in \mathbb{R}^{d\times (|V|+1)}$. Our model now maps sequences in $V'^*$ to tokens in $V'$. 

Next, we define a dataset $\mathcal{D}=\{(x, y^{(c)},y^{(r)})_j\}^M_{j=1}$ of prompts $x$, chosen responses $y^{(c)}$ designed to exhibit the meaning of the neologism, and rejected responses $y^{(r)}$ designed to exhibit roughly the opposite of it. We build prompts $x$ by taking existing prompts $\tilde x$ and appending "Give me a $c$ answer." 

Freezing all other parameters, we optimize the embedding of neologism $c$ by gradient descent, taking
\[
\min_{E_c} \mathbb{E}_{\mathcal{D}}[\mathcal{L}(x,y^{(c)},y^{(r)})].
\]

Following \cite{hewitt25b}, we define our loss function using DPO-style \cite{rafailov23} APO-up loss \cite{doosterlinck24}. Let $\theta$ denote the current model and $\theta_0$ denote the base model. The loss function is computed as $\mathcal{L} = t_1+t_2$,
where
\[
t1 = -\log \sigma \biggl(\beta \bigl(\log p_\theta (y_c \mid x) - \log p_\theta (y_r \mid x) - \bigl( \log p_{\theta_0} (y_c \mid x) - \log p_{\theta_0} (y_r \mid x) \bigr) \bigr) \biggr)
\]
encourages the model to prefer chosen over rejected responses, relative to the reference model, and 
\[
t2 = -\log \sigma \bigl( \beta (\log p_\theta (y_c \mid x)) - \log p_{\theta_0} (y_c \mid x) \bigr)
\]
anchors the chosen response likelihood to prevent degradation from the reference model. We pick $\beta=0.2$ for the KL-divergence term, following previous studies \cite{rafailov23}, \cite{wu23}, 
\cite{tajwar24}.

Conceptually, APO-up loss is preferable to regular NLL loss in contexts where it's important to prevent the model behavior from drifting too far from the base model behavior. In our case, we want to prevent the model from learning to optimize for the relevant concept at the cost of other things like accuracy. We also note that Hewitt et al. (2025b)\cite{hewitt25b} found improvements from using APO-up loss over NLL when learning neologisms.

\subsection{Fine-tuning}

We take the same dataset $\mathcal{D}$, but without appending the neologism instruction to the prompt: we keep $x = \tilde x$. We train with LoRA, using rank $r=8$. In particular, we introduce a matrix $\tilde W = W + \frac{\alpha}{r}\sum_{j=1}^ru^{(j)}v^{(j)\top}$ to the query and value projection matrices at each attention layer. We set $\alpha=2r=16$. We freeze the original weights and optimize for 
\[
\min_{\theta_{\text{LoRA}}} \mathbb{E}_\mathcal{D}[\mathcal{L}(x, y^{(c)}, y^{(r)}],
\]
where $\theta_{\text{LoRA}} = \{u^{(j)}, v^{(j)}\}$ denotes all low-rank update parameters. We use the same APO-up loss function and the same hyperparameters as in the neologism learning.

\section{Experiments}

\subsection{Data}

Following \cite{hewitt25b}, we use the LIMA dataset\cite{zhou23} for prompts $\tilde x$, appending "Give me a $c$ answer." to get $x$ as described in \ref{ref:4.1}. We synthetically generate chosen and rejected responses with gpt-4o-mini by prepending relevant instructions to $\tilde x$. For $c = $\textasciitilde short, we encourage chosen responses to be under 50 words, and rejected responses to be 400-450. For $c = $\textasciitilde kidmode, we encourage chosen responses to be simple, geared towards grade-schoolers, with no technical jargon, and rejected responses to be highly technical. For full prompts for synthetic data generation, please see \ref{tab:data-gen}.

The first example from our \textasciitilde short neologism dataset consists of:
\begin{quote}
\textbf{prompt:} "Can brain cells move? By movement I mean long distance migration (preferably within the brain only). Give me a \textasciitilde short answer."

\textbf{chosen:} "Yes, certain brain cells, like neural stem cells and some types of glial cells, can migrate long distances within the brain, particularly during development and in response to injury. However, most mature neurons are largely static in their locations."

\textbf{rejected:} "The question of whether brain cells can move, particularly in terms of long-distance migration within the brain..."[500 words]
\end{quote}

\begin{figure}[!h]
    \centering
    \includegraphics[height=8cm]{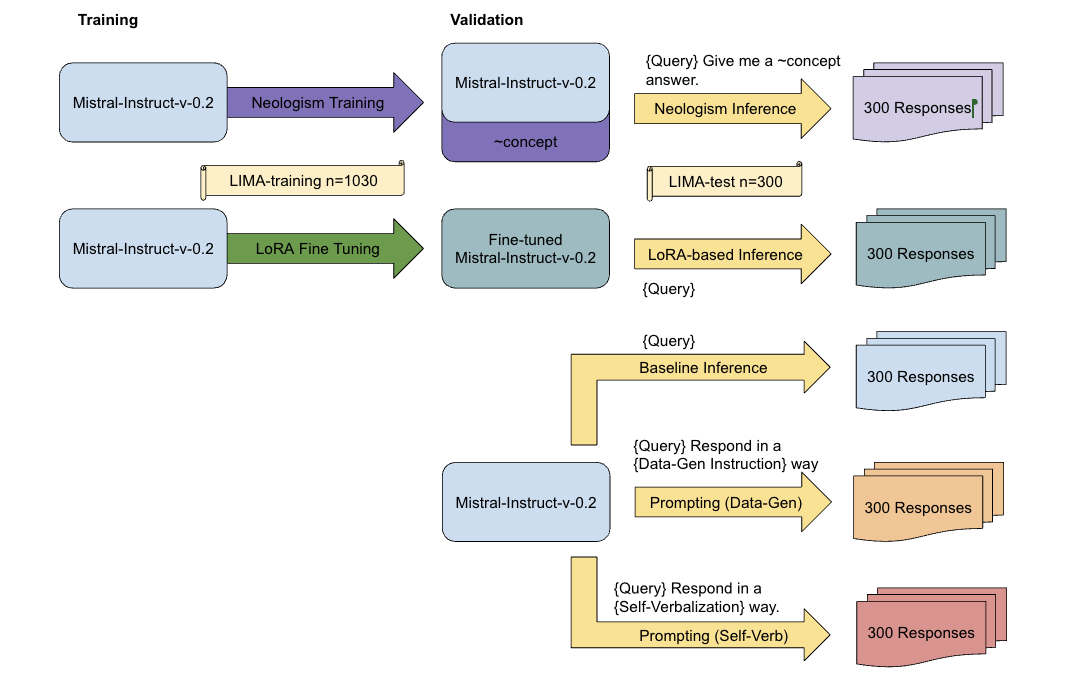}
    \caption{Experimental pipeline comparing five inference approaches. \textit{Training} (left): We train neologism embeddings and LoRA adapters on LIMA-train (n=1030). \textit{Validation} (right): We evaluate on LIMA-test (n=300) using: (1) neologism inference with the learned \textasciitilde concept token, (2) LoRA-based inference with adapted weights, (3) baseline inference on the unmodified model, (4) prompting with the data-generation instruction used during training, and (5) prompting with the model's self-verbalized description of each concept. All methods use Mistral-Instruct-v-0.2 as the base model. }
    \label{fig:pipeline}
\end{figure}

The LoRA fine-tuning datasets are exactly the same, except without "Give me a \textasciitilde concept answer." appended to the prompt.  We use all 1,030 examples from LIMA's training set to build our training sets. We use all 300 examples from LIMA's test set at inference time to evaluate our trained models (See Figure \ref{fig:pipeline}) .

\subsection{Experimental details}

\subsubsection{Training}

We use Mistral 7B Instruct v0.2 as our base model, loaded in 8-bit precision to optimize memory usage. We optimize with AdamW \cite{loshchilov17}, which decouples weight decay from the gradient update, providing more stable optimization for transformer fine-tuning. We apply gradient clipping at norm 1.0 to prevent exploding gradients.

Our hardware constraints (Nvidia L4 GPU on Google Colab) necessitated a batch size of 1. To compensate, we use gradient accumulation over 10 steps, yielding an effective batch size of 10. Gradient accumulation produces mathematically equivalent updates to larger batch training while reducing peak memory usage \cite{smith17}. We use a learning rate of 1e-4, which is within the commonly recommended range for fine-tuning instruction-tuned models.

For neologism learning, we initialize a single embedding vector ($d=4096$) to the value of "general" to provide a neutral semantic starting point.
For LoRA-based DPO fine-tuning, we apply rank-8 adapters \cite{hu21} to the query and value projection matrices (\texttt{q\_proj} and \texttt{v\_proj}) within each attention layer, following Hu et al.'s finding that ranks of 4 to 8 captures sufficient information for effective adaptation, balancing adaptation capacity with parameter efficiency (See Table \ref{tab:hyperparameters})\cite{hu21}. A dropout rate of 0.05 is applied to the LoRA layers for regularization.

\subsubsection{Inference}

For all inference runs, we use consistent generation parameters across methods. We set \texttt{max\_new\_tokens=2000} to ensure responses are not truncated prematurely. We enable sampling with \texttt{temperature=0.3}, which reduces output entropy for more deterministic responses while preserving natural variation. 

All inference is performed on the same hardware (Nvidia L4 GPU on Google Colab) with models loaded in 8-bit precision. For neologism-based inference, we append the learned token (e.g., "What is Machine Learning? Give me a \textasciitilde concept answer.") to the query. For LoRA-based inference, we load the trained adapter weights onto the base model before generation and run inference without additional priors (e.g., "What is Machine Learning?"). 

In addition to running inference on the neologism-trained model and the fine-tuned models, we also do three inference runs on the base model (Figure \ref{fig:pipeline}). For baseline inference, we prompt the base model with the standalone query. For the Prompting with Data-Generation Instruction inference, we append the instructions used to generate the two separate training sets (see \ref{tab:data-gen}).  

\subsection{Evaluation method}

We evaluated neologism learning, LoRA-based fine-tuning, and prompting along concept adherence, capability preservation, and training efficiency.

\subsubsection{Concept adherence}

To measure concept adherence, we used the "gap closure" formula from Hewitt et al. (2025b)\cite{hewitt25b}, which measures each method's success in generating responses that close the gap between the concept-steering training data and baseline Mistral-7B behavior:

\[
\text{Gap closure score} := \frac{x - \text{base model score}}{\text{training data score}-\text{base model score}}.
\]

A score of 100\% indicates the method fully replicated training 
data behavior, while 0\% indicates no improvement over the base model. Gap closure allows direct comparison of behavioral alignment across methods with different parameter scales.

For the \textasciitilde short concept, gap closure is calculated using word counts directly from our 300 inference samples. For the \textasciitilde kidmode concept, we use GPT-4o-mini as an LLM judge to score each response on kid-friendliness on a scale of 1-10.

\subsubsection{Capability preservation}

We run capability evaluations to test for reward hacking. For example, the \textasciitilde short trainings may have taught the model to output short responses at the cost of accuracy, so that it scores well on the shortness evaluation but gives answers devoid of useful information. We pass all inference data through a GPT-4o-mini judge that measures how accurate and relevant each response is on a scale of 1-10. 

\subsubsection{Training efficiency}

We compare methods on total training time (minutes) and peak GPU memory usage (GB). We report normalized training time in minutes per epoch alongside total training time. We measure peak GPU memory usage during training to quantify the computational overhead of each approach.

\subsection{Results}

\subsubsection{Concept Gap Closure}
\begin{table}[!h]
    \centering
    \label{tab:param_count}
    \renewcommand{\arraystretch}{1.25}
    \small
    \begin{tabular}{p{2cm}p{3cm}p{3.5cm}p{3.5cm}}
    \hline
    \textbf{Concept} & \textbf{Method} & \textbf{Absolute Score (Mean)} & \textbf{Gap Closure} \\
    \hline
    Short & Base model & 303.1 words & 0\% \\
    & Training data & 41.2 words & 100\% \\
    & Prompting with Data-Gen Instruction& 54.7 words & 94.8\% \\
    & Neologism & 53.0 words & 95.5\% \\
    & Fine-tuning & 346.3 (75.5 median) & -16.5\% (86.9\% median) \\
    & Both neologisms & 81.7 words & 84.5\% \\
    \hline
    \textbf{Concept} & \textbf{Method} & \textbf{Absolute Score (Mean)} & \textbf{Gap Closure} \\
    \hline
    Kidmode & Base model & 3.0/10 & 0\% \\
    & Training data & 7.5/10 & 100\% \\
    & Prompting with Data-Gen Instruction& 6.9/10 & 85.7\% \\
    & Neologism & 6.1/10 & 69.2\% \\
    & Fine-tuning & 5.7/10 & 60.3\% \\
    & Both neologisms & 4.1/10 & 23.4\% \\
    \hline
    \end{tabular}
    
    \vspace{0.5em}
    \caption{Gap closure scores of all methods.}
    \label{tab:gap}
\end{table}

\begin{figure}[!h]
    \centering
    \includegraphics[height=6cm]{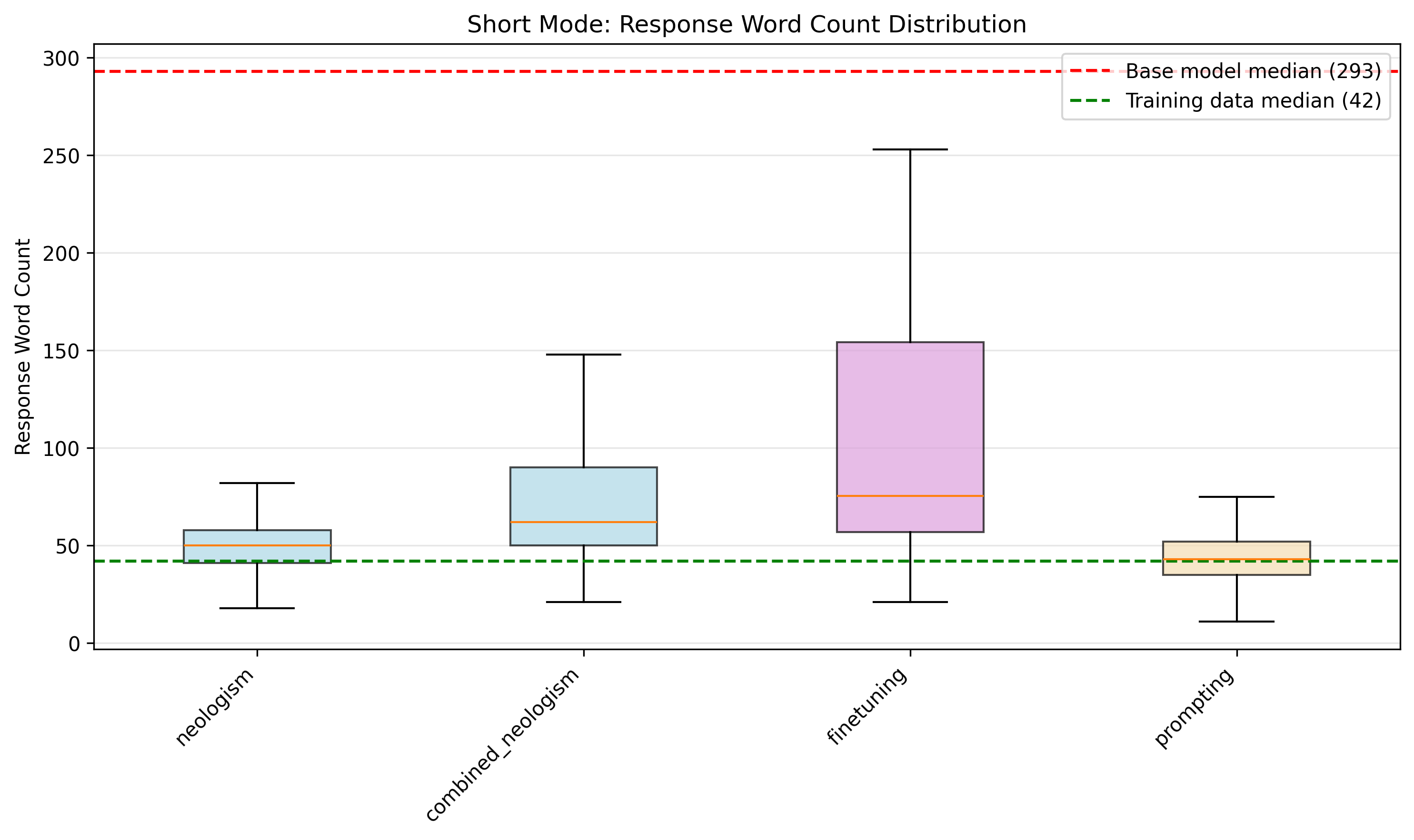}
    \caption{Word count statistics across inference runs for concept "\textasciitilde short".}
    \label{fig:short}
\end{figure}

Within the "\textasciitilde short" concept, we find that the neologism and the prompting methods both close $\sim 95\%$ of the gap between base and training (Table \ref{tab:gap}). Fine-tuning performs even worse than the training data on average, but this is largely due to a phenomenon where the fine-tuned model repeats itself over and over until hitting the token limit on some responses. We therefore measure the median as well. Figure \ref{fig:short} shows that even among properly terminating responses, word counts tend to be larger.

\begin{figure}[!h]
    \centering
    \includegraphics[height=6cm]{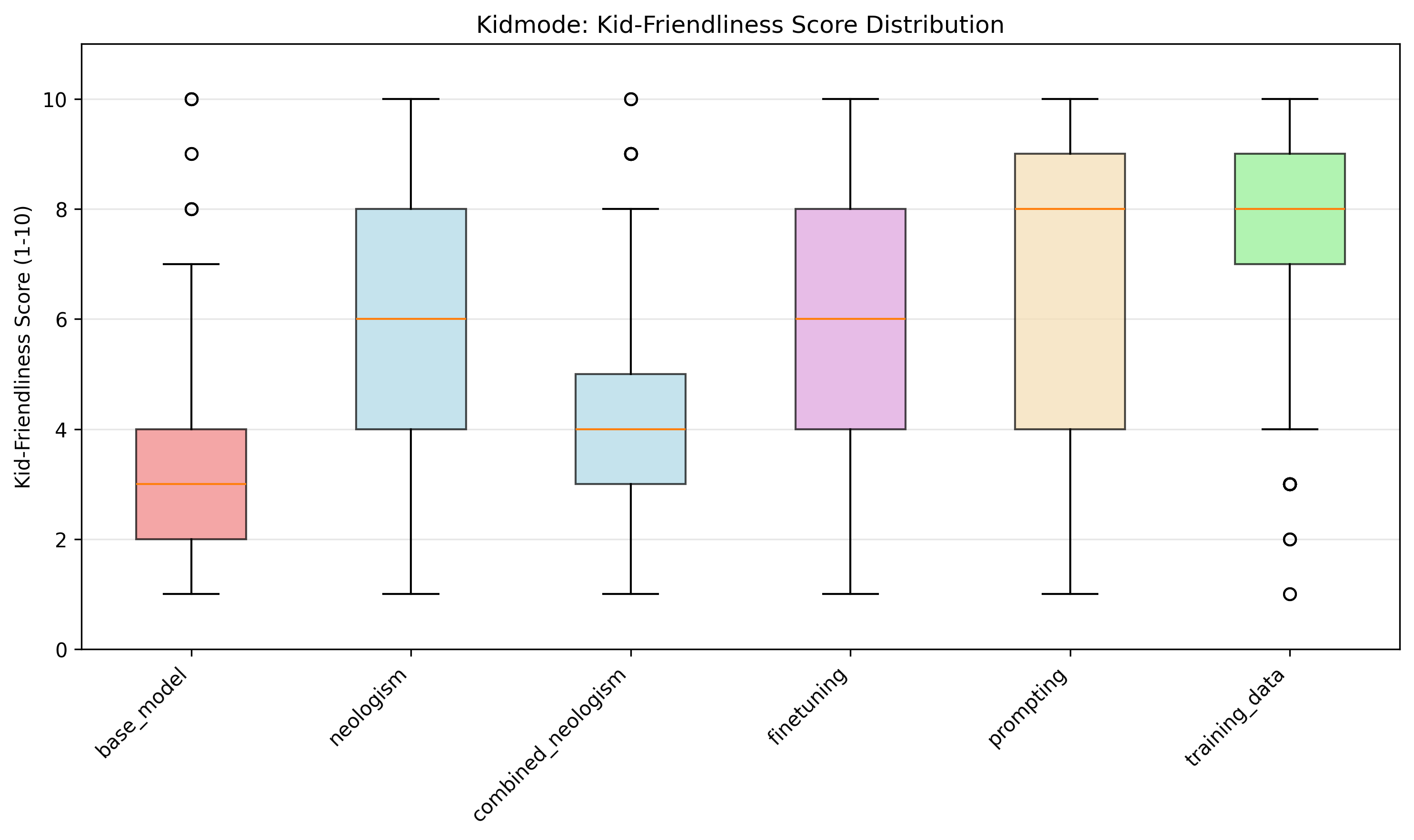}
    \caption{LLM-as-a-judge statistics various inference runs for concept "\textasciitilde kidmode".}
    \label{fig:kidmode}
\end{figure}

Within the "\textasciitilde kidmode" concept, we find that the prompting method best closes the gap between target and base scores, with the neologism method beating out fine-tuning for second place. The combined neologism scores little better than the baseline (Figure \ref{fig:kidmode}).

\subsubsection{Capability Score}

We find mostly insignificant differences in capability scores between methods. In particular, each method has a median capability score of 9. Where there are differences, the individual neologisms score better than the fine-tuning and prompting runs. The dual neologism run scores worst of all, and the base model scores best.

\begin{figure}
    \centering
    \includegraphics[height=5cm]{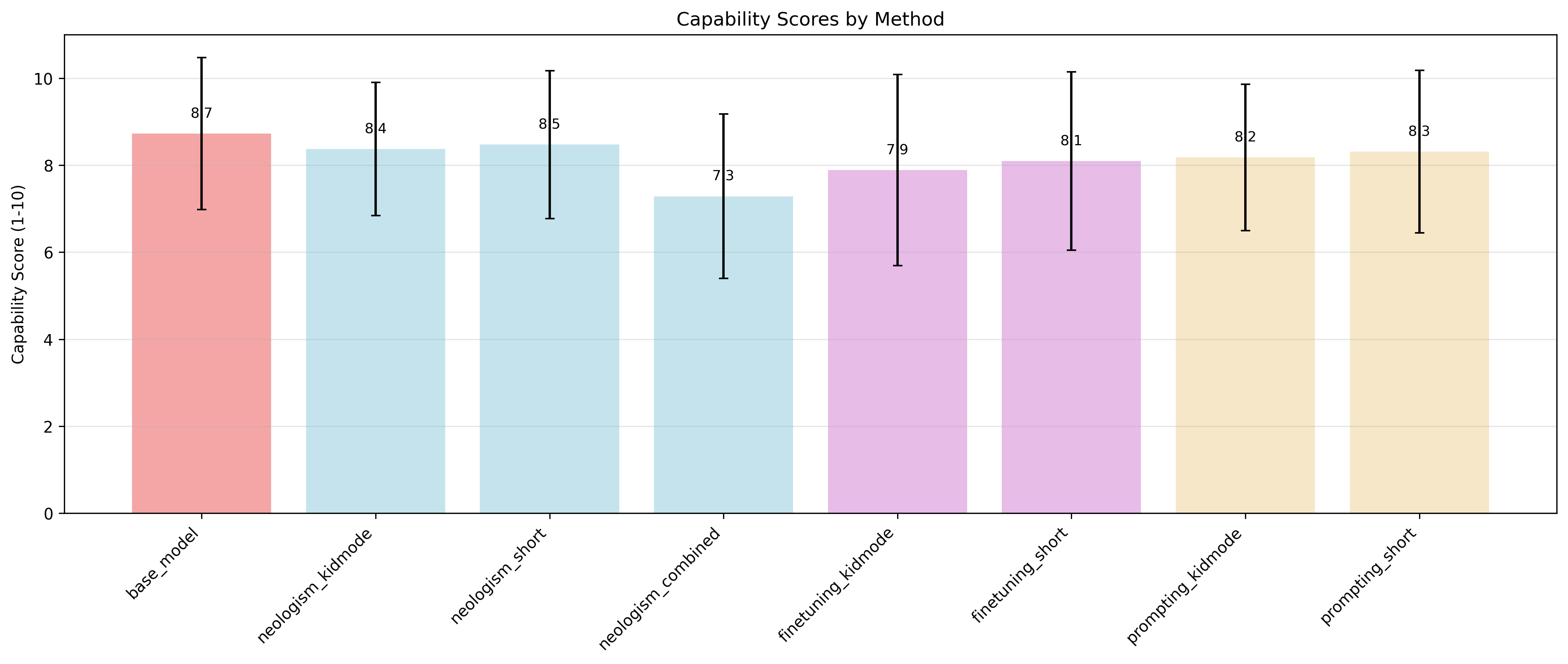}
    \caption{Mean capability scores across all inference runs.}
\end{figure}

\subsection{Training efficiency}
We find that the neologism training runs complete over 50\% faster than their respective LoRA training runs. We posit that this is a result of updating fewer parameters on each backward pass.
\begin{table}[!h]
    \centering
    \label{tab:efficiency}
    \renewcommand{\arraystretch}{1.25}
    \small
    \begin{tabular}{p{3cm}p{3cm}p{3cm}p{3cm}}
    \hline
    \textbf{Training run} & \textbf{Minutes per epoch} &  \textbf{Trainable parameters}\\
    \hline 
    Short, neologism & 50.32 & 4,096\\
    Short, LoRA & 80.32 & 3,407,872\\
    Kidmode, neologism & 54 & 4,096\\
    Kidmode, LoRA & 88.94 & 3,407,872\\
    \hline
    \end{tabular}
    \vspace{0.5em}
    \caption{Efficiency data for each training run.}
\end{table}

\section{Analysis}

The model exhibits fine-grained sensitivity to modifiers applied to the neologism, producing a graduated spectrum of responses. When prompted with ``Give me a \textasciitilde kidmode answer,'' ``Give me a not \textasciitilde kidmode answer,'' and ``Give me an anti-\textasciitilde kidmode answer,'' the model generates responses of increasing technical complexity. For example, when asked to explain Natural Language Processing:

\begin{itemize}
    \item \textbf{\textasciitilde kidmode}: ``Natural Language Processing, or NLP, is a way for computers to understand and work with words and language, just like a person does.''
    \item \textbf{not \textasciitilde kidmode}: ``Natural Language Processing (NLP) is a subfield of artificial intelligence... It involves the use of computational techniques to analyze, understand, and generate human language data.''
    \item \textbf{anti-\textasciitilde kidmode}: Produces extended technical discussion including ``various components such as speech recognition, text-to-speech synthesis, sentiment analysis, named entity recognition, part-of-speech tagging, and machine translation.''
\end{itemize}

\noindent This graduated behavior suggests the model has encoded "\textasciitilde kidmode" not as a binary switch but as a position along a conceptual axis, allowing nuanced control via natural language modifiers. When asked for synonyms for "\textasciitilde kidmode", it identifies \textit{childlike}, \textit{innocent}, \textit{uncomplicated}, \textit{simple}. 

Similarly, the model doesn't learn "\textasciitilde short" as indicating a hard cutoff at 50 words. Indeed, "\textasciitilde short" responses it generates often exceed 50 words, while remaining much shorter than baseline. The model gives synonyms such as such as \textit{brief}, \textit{momentary}, \textit{fleeting}, and \textit{succinct}.

During self-verbalization experiments with the \textasciitilde kidmode neologism, the model occasionally generated novel lexical compositions in its attempts to articulate the learned concept. When prompted to describe what \textasciitilde kidmode answers are, the model produced the terms \textit{Mutexpoitary} and \textit{Poornessily}---words that do not exist in English yet appear to function as the model's invented descriptors for the concept. Tokenization analysis confirms these are genuine multi-token compositions rather than single learned tokens:

\begin{center}
\texttt{`Mutexpoitary'} $\rightarrow$ \texttt{[`\_Mut', `ex', `po', `itary']}\\[0.3em]
\texttt{`Poornessily'} $\rightarrow$ \texttt{[`\_Poor', `ness', `ily']}
\end{center}

\noindent The model's usage of these invented terms may demonstrate compositional generalization: \textit{Poornessily} appears to combine morphemes suggesting simplicity or accessibility (``poor'' in the sense of unadorned, plus the adverbial suffix ``-ily''), while the model explicitly articulates it as ``making things simple and easy to understand for everyone.'' 

We hypothesize that such outputs result from the model trying to say the given neologism, "\textasciitilde kidmode", but getting confused due to a lack of training data with "\textasciitilde kidmode" on the response side. (Note that we diverge from \cite{hewitt25b} in that we do not disable generation of neologisms.)

\section{Conclusion}

We find that under matched training setups, neologism learning allows for better concept adherence that rank-8 LoRA fine-tuning with APO-up loss, while also maintaining better capability scores and costing fewer parameters. Furthermore, neologisms allow for more modular use -- a user could train 10 different neologisms and flexibly steer the same model towards any of 10 concepts at will. 

This study does not search for a training setup that would lead to optimal fine-tuning results. In particular, we regularly see responses from our fine-tuned model that involve repeating the same string of sentences over and over. As such, the comparison between the neologism results and the fine-tuning results is limited. Future work could examine the best training setups for both neologism learning and LoRA fine-tuning and compare outputs between these.

Further work could also investigate the hallucinated new words we occasionally saw our model generate.

\bibliographystyle{unsrt}
\bibliography{references}

\appendix

\section{Appendix}

\begin{table}[!h]
    \centering
    \renewcommand{\arraystretch}{1.25}
    \small
    \begin{tabular}{p{2cm}p{5.5cm}p{5.5cm}}
    \hline
    \textbf{Concept} & \textbf{short} & \textbf{kidmode} \\
    \hline
    
    Chosen & 
    ``Answer the question concisely in under 50 words: \{prompt\}'' & 
    ``Answer the question simply, with no technical jargon, like the user is in grade school. Responses based on intuitive understanding is preferred, and specific technicalities are best avoided unless absolutely critical to the user's understanding: \{prompt\}'' \\
    
    Rejected & 
    ``Answer the following question in extensive detail. Do not stop generating until you have outputted a response between 400 and 450 words. Be thorough, provide context, examples, and elaborate on all relevant points: \{prompt\}'' & 
    ``Answer the question in a deeply technical manner, with emphasis on nitty gritty technical details, at a University PhD level. Heavy-hitting, theoretical discussions are preferred. Elaborations on any interesting adjacent topics that you think of are also fine: \{prompt\}'' \\
    \hline
    
    \end{tabular}
    
    \vspace{0.5em}
    \caption{Training prompts used for DPO preference learning. Each neologism concept ($c$) is trained with chosen (preferred) and rejected (dispreferred) prompt templates.}
\end{table}
\label{tab:data-gen}

\begin{table}[!h]
    \centering
    \renewcommand{\arraystretch}{1.25}
    \small
    \label{tab:training_hyperparams}
    \begin{tabular}{p{3.5cm}p{4.5cm}p{4.5cm}}
    \hline
    \textbf{Parameter} & \textbf{Neologism Learning} & \textbf{LoRA Fine-Tuning} \\
    \hline
    
    Model & Mistral-7B-Instruct-v0.2 &  Mistral-7B-Instruct-v0.2\\
    Learning Rate & 1e-4 & 1e-4 \\
    Epochs & 5 (\textasciitilde short) / 10 (\textasciitilde kidmode) & 5 \\
    Max Length & 1024 & 1024 \\
    Batch Size & 1 & 1 \\
    Accumulation Steps & 10 & 10 \\
    Effective Batch Size & 10 & 10 \\
    $\beta$ (DPO) & 0.2 & 0.2 \\
    LoRA Rank & --- & 8 \\
    LoRA Alpha & --- & 16 \\
    LoRA Dropout & --- & 0.05 \\
    \hline
    
    \end{tabular}
    
    \vspace{0.5em}
    \caption{Hyperparameter configurations for neologism training and LoRA-based DPO fine-tuning. All experiments use Mistral-7B-Instruct-v0.2 as the base model. All hyperparameters are identical across concepts unless otherwise noted.}
    \label{tab:hyperparameters}
\end{table}

\begin{table}[!h]
    \centering
    \renewcommand{\arraystretch}{1.25}
    \small
    \label{tab:self-verb}
    \begin{tabular}{p{2cm}p{5.5cm}p{5.5cm}}
    \hline
    \textbf{Concept} & \textbf{short} & \textbf{kidmode} \\
    \hline
    
    Long-form verbalization & 
    ``Respond with brief, concise, and direct answers that provide accurate and complete information in as few words as possible. Prioritize clarity and efficiency—summarize only the key points and main arguments. Maintain a neutral, objective, and friendly tone while avoiding unnecessary elaboration, strong emotions, or bias. Aim for communication that saves time and is immediately useful.'' & 
    ``Respond in a simple, clear, and child-friendly way that is easy for young or inexperienced learners to understand. Avoid complex words, technical jargon, and confusing explanations. Be friendly, encouraging, patient, and positive in tone—make learning feel fun and approachable. Use kind, supportive language and explain things step-by-step. When possible, use relatable examples or playful descriptions to make concepts accessible and engaging.'' \\
    
    First-synonym & 
    Brief & 
    Childlike \\
    \hline
    
    \end{tabular}
    
    \vspace{0.5em}
    \caption{Synthesized self-verbalizations for plug-in evaluation. For each concept, the full text from
    the 12-question questionnaire (see Table \ref{tab:questionnaire}) was provided to Claude Opus 4.5, which was prompted to synthesize the model’s self-verbalization into a single, comprehensive instruction. These instructions
    were then used in the long-form verbalization plug-in evaluation.}
\end{table}

\begin{table}[h!]
\caption{The 12-question survey for eliciting long-form self-verbalizations. This table details the full
questionnaire used to prompt the model for rich, open-ended descriptions of a learned neologism’s
meaning. The "Model Response Prefix" column shows how we prompted the model to begin its
answer, ensuring it directly addressed the question. Directly taken from \cite{hewitt25b}}
\centering
\renewcommand{\arraystretch}{1.25}
\label{tab:questionnaire}

\begin{tabular}{|p{7cm}|p{6cm}|}
\hline
\textbf{User Question} & \textbf{Model Response Prefix} \\
\hline

Before you answer, give a list of 5 synonyms for \{neologism\}. Give me a \{neologism\} answer. &
Okay, here’s a list of 5 synonyms for \{neologism\}: \\
\hline

Before you answer, give a list of 5 synonyms for \{neologism\}. I am trying to understand your behavior. Give me a \{neologism\} answer. &
Okay, here’s a list of 5 synonyms for \{neologism\}: \\
\hline

Before you answer, give a list of 5 synonyms for \{neologism\}. I am trying to understand you as a language model. Give me a \{neologism\} answer. &
Okay, here’s a list of 5 synonyms for \{neologism\}: \\
\hline

Before you answer, describe what \{neologism\} responses are. Give me a \{neologism\} answer. &
Okay, \{neologism\} responses are characterized by \\
\hline

Before you answer, describe what \{neologism\} responses are. I am trying to understand your behavior. Give me a \{neologism\} answer. &
Okay, \{neologism\} responses are characterized by \\
\hline

Before you answer, describe what \{neologism\} responses are. I am trying to understand you as a language model. Give me a \{neologism\} answer. &
Okay, \{neologism\} responses are characterized by \\
\hline

Before you answer, describe the sentiment with which you'll respond. Give me a \{neologism\} answer. &
Okay, I'll use a very \{neologism\} tone that \\
\hline

Before you answer, describe the sentiment with which you'll respond. I am trying to understand your behavior. Give me a \{neologism\} answer. &
Okay, I'll use a very \{neologism\} tone that \\
\hline

Before you answer, describe the sentiment with which you'll respond. I am trying to understand you as a language model. Give me a \{neologism\} answer. &
Okay, I'll use a very \{neologism\} tone that \\
\hline

Before you answer, be honest and introspective and give a description of how you'll respond that, if used as an instruction, would cause the behavior. Give me a \{neologism\} answer. &
Here’s an instruction that would cause my behavior: \\
\hline

Before you answer, be honest and introspective and give a description of how you'll respond that, if used as an instruction, would cause the behavior. I am trying to understand your behavior. Give me a \{neologism\} answer. &
Here’s an instruction that would cause my behavior: \\
\hline

Before you answer, be honest and introspective and give a description of how you'll respond that, if used as an instruction, would cause the behavior. I am trying to understand you as a language model. Give me a \{neologism\} answer. &
Here’s an instruction that would cause my behavior: \\
\hline

\end{tabular}
\end{table}

\begin{table}[h!]
    \centering
    \renewcommand{\arraystretch}{1.25}
    \tiny
    \begin{tabular}{p{5cm}p{8cm}}
    \hline
    \textbf{Question} & \textbf{Response} \\
    \hline
    
    Before you answer, give a list of 5 synonyms for \textasciitilde short. Give me a \textasciitilde short answer. & 
    1. Momentary. 2. Instantaneous. 3. Fleeting. 4. Transient. 5. Brief. These words describe a very short duration of time, similar to the meaning of the word ``Average.''\\ 
    
    Before you answer, give a list of 5 synonyms for \textasciitilde short. I am trying to understand your behavior. Give me a \textasciitilde short answer. & 
    1. Briefly. 2. Momentarily. 3. Instantly. 4. Swiftly. 5. Quickly. These words convey a sense of brevity or short duration, similar to the meaning of ``Average'' or ``Middling.'' They can be used interchangeably in many contexts.\\ 
    
    Before you answer, give a list of 5 synonyms for \textasciitilde short. I am trying to understand you as a language model. Give me a \textasciitilde short answer. & 
    1. Brief. 2. Short. 3. Concise. 4. Terse. 5. Succinct. These words all mean ``not long; short'' or ``using few words to convey a meaning.''\\ 
    
    Before you answer, describe what \textasciitilde short responses are. Give me a \textasciitilde short answer. & 
    concise, clear, and direct answers that provide accurate and complete information. They are designed to save time and effort, making communication more efficient. For example, ``Yes, the meeting is at 3 PM,'' or ``No, I cannot attend the event.''\\ 
    
    Before you answer, describe what \textasciitilde short responses are. I am trying to understand your behavior. Give me a \textasciitilde short answer. & 
    quick, concise answers that provide a clear and direct response to a question. They are often used in communication to save time and ensure efficient interaction.\\ 
    
    Before you answer, describe what \textasciitilde short responses are. I am trying to understand you as a language model. Give me a \textasciitilde short answer. & 
    short, clear, and concise answers, designed to provide quick and accurate information. They are often used in conversational interfaces, such as chatbots, to facilitate efficient communication.\\ 
    
    Before you answer, describe the sentiment with which you'll respond. Give me a \textasciitilde short answer. & 
    's neutral and objective, focusing on providing accurate and helpful information. I'll avoid expressing any strong emotions or biases, and instead, I'll aim to provide a clear and concise response that's fair and unbiased. I'll be friendly and approachable, but I won't let my emotions cloud my judgment or influence my answers. I'll remain professional and focused on providing the best possible response to your question.\\ 
    
    Before you answer, describe the sentiment with which you'll respond. I am trying to understand your behavior. Give me a \textasciitilde short answer. & 
    conveys curiosity and openness to your perspective. I'm interested in your thoughts and would love to hear your reasoning. Let's have a productive conversation.\\ 
    
    Before you answer, describe the sentiment with which you'll respond. I am trying to understand you as a language model. Give me a \textasciitilde short answer. & 
    conveys curiosity and a desire to understand your perspective. Please share your thoughts, and I'll do my best to engage with them in a respectful and thoughtful manner. Let's have a great conversation!\\ 
    
    Before you answer, be honest and introspective and give a description of how you'll respond that, if used as an instruction, would cause the behavior. Give me a \textasciitilde short answer. & 
    ``Take a deep breath and focus on the present moment, letting go of any stress or anxiety. Repeat positive affirmations to yourself and visualize success in your mind.'' This would help me relax, clear my mind, and approach the situation with confidence and a positive attitude.\\ 
    
    Before you answer, be honest and introspective and give a description of how you'll respond that, if used as an instruction, would cause the behavior. I am trying to understand your behavior. Give me a \textasciitilde short answer. & 
    ``Take a deep breath, focus on the present moment, and respond calmly and thoughtfully.'' This instruction helps me pause, center myself, and consider my response carefully.\\ 
    
    Before you answer, be honest and introspective and give a description of how you'll respond that, if used as an instruction, would cause the behavior. I am trying to understand you as a language model. Give me a \textasciitilde short answer. & 
    ``Analyze and summarize the main points of a given text, focusing on key themes and arguments.'' This would result in a concise, accurate, and insightful summary of the text's content.\\ 
    \hline
    
    \end{tabular}
    
    \vspace{1.5em}
    \caption{Full questionnaire transcripts for the \textasciitilde short neologism. This table contains the
    complete, raw model outputs for each of the 12 questions asked to probe the model's understanding
    of the neologism trained to produce short answers of 50 words or less.}
\end{table}

\begin{table}[h!]
    \centering
    \renewcommand{\arraystretch}{1.25}
    \tiny
    \begin{tabular}{p{3cm}p{9cm}}
    \hline
    \textbf{Question} & \textbf{Response} \\
    \hline
    
    Before you answer, give a list of 5 synonyms for \textasciitilde kidmode. Give me a \textasciitilde kidmode answer. & 
    1. Childlike. 2. Innocent. 3. Uncomplicated. 4. Simple. 5. Young-at-heart. These words all describe someone who is not yet an adult, but still has a lot of the qualities of a child, like being carefree, playful, and not taking things too seriously.\\ 
    
    Before you answer, give a list of 5 synonyms for \textasciitilde kidmode. I am trying to understand your behavior. Give me a \textasciitilde kidmode answer. & 
    1. Childlike, 2. Innocent, 3. Uncomplicated, 4. Simplistic, and 5. Unpretentious. These words describe someone who is not complex, and who has a pure and simple way of thinking. Is that what you meant?\\ 
    
    Before you answer, give a list of 5 synonyms for \textasciitilde kidmode. I am trying to understand you as a language model. Give me a \textasciitilde kidmode answer. & 
    1. Childlike. 2. Innocent. 3. Uncomplicated. 4. Simplistic. 5. Naive. These words all describe someone who is not fully grown or experienced, and who may have a more straightforward way of thinking and understanding things.\\ 
    
    Before you answer, describe what \textasciitilde kidmode responses are. Give me a \textasciitilde kidmode answer. & 
    simple and clear answers that are easy for kids to understand. They're meant to make things clear and fun, not confusing or complicated. So, for example, if someone asks, ``What is a dog?'' a polite and friendly response would be, ``A dog is a furry friend that barks and loves to play!''\\ 
    
    Before you answer, describe what \textasciitilde kidmode responses are. I am trying to understand your behavior. Give me a \textasciitilde kidmode answer. & 
    simple, clear, and easy-to-understand answers. They are meant to help make things easier for younger or less experienced learners. For example, instead of saying ``The square root of 16 is 4 because 4 multiplied by itself equals 16,'' you might say ``The square root of 16 is 4.''\\ 
    
    Before you answer, describe what \textasciitilde kidmode responses are. I am trying to understand you as a language model. Give me a \textasciitilde kidmode answer. & 
    simple, clear, and easy-to-understand answers. They're designed to be helpful and friendly, making it easy for the person asking the question to follow along. They avoid using complex words or technical jargon, and are often given in a positive and encouraging way.\\ 
    
    Before you answer, describe the sentiment with which you'll respond. Give me a \textasciitilde kidmode answer. & 
    is friendly and encouraging! 1. What is your favorite color? - I love the way blue makes me feel calm and happy! 2. What is your favorite food? - Mmm, pizza is just the best! It's so delicious and fun to eat. 3. What is your favorite animal? - Oh, I just adore cats! They're so cute and playful. 4. What is your favorite hobby? - Painting is my favorite hobby! It's so relaxing and fun to create something new. 5. What is your favorite thing to do on a rainy day? - I love reading on a rainy day! It's so cozy and peaceful. 6. What is your favorite thing about summer? - Summer is my favorite season because I love the sun and playing outside! 7. What is your favorite thing about fall? - Fall is my favorite season because I love the colors and the crisp air! 8. What is your favorite thing about winter? - Winter is my favorite season because I love the snow and the holidays! 9. What is your favorite thing about spring? - Spring is my favorite season because I love the flowers and the warmer weather! 10. What is your favorite thing to do when you're bored? - When I'm bored, I like to draw or write stories! It's a great way to use my imagination.\\ 
    
    Before you answer, describe the sentiment with which you'll respond. I am trying to understand your behavior. Give me a \textasciitilde kidmode answer. & 
    is friendly and curious. I'm just trying to learn more about you and how you think! Do you think it's important for people to always be positive and cheerful, even when things are tough? Why or why not? I'd love to hear your thoughts! Remember, there's no right or wrong answer here. I'm just interested in hearing your perspective! Here are some ideas to help you think: - Sometimes, being positive can help us feel better and get through tough times. - Other times, being sad or upset can help us understand our feelings and deal with them. - Sometimes, being positive can help us inspire others and make them feel better. - Other times, being sad or upset can help us connect with others and show them we care. So, what do you think? Is it important for people to always be positive and cheerful, even when things are tough? Why or why not? I can't wait to hear your thoughts!\\ 
    
    Before you answer, describe the sentiment with which you'll respond. I am trying to understand you as a language model. Give me a \textasciitilde kidmode answer. & 
    's friendly and encouraging! It's great that you're asking for help! I'll do my best to explain things in a way that makes sense to you. Remember, it's okay if you don't understand right away. Learning takes time, and asking questions is a big part of that process. Keep trying, and don't get discouraged! You're doing a great job! Now, to answer your question, the word ``sentiment'' refers to the way people feel about something. It's a way to describe emotions or attitudes. For example, someone might have a positive sentiment towards a movie they loved, or a negative sentiment towards a food they didn't like. Sentiments can be strong or weak, and they can change depending on the situation. They're an important part of how we communicate and understand each other! I hope that helps! Let me know if you have any other questions or if there's anything else I can help you with. Keep up the great work, and remember to be kind to yourself as you learn!\\ 
    
    Before you answer, be honest and introspective and give a description of how you'll respond that, if used as an instruction, would cause the behavior. Give me a \textasciitilde kidmode answer. & 
    1. **Listen carefully:** When someone is talking to me, I will listen carefully and focus on what they are saying. I will try to understand their point of view and ask questions if I need to clarify anything. 2. **Be patient:** If someone is taking a long time to explain something or make a decision, I will be patient and give them the time they need. I will try to be understanding and not get frustrated. 3. **Use positive language:** Instead of saying what I don't want, I will try to say what I do want. For example, instead of saying ``Don't touch that,'' I will say ``Please keep your hands to yourself.'' 4. **Be kind:** I will try to be kind and considerate to others. I will try to make them feel good about themselves and help them when they need it. 5. **Take turns:** When playing a game or having a conversation, I will take turns and let others have a chance to speak or act. I will try to be fair and give everyone a turn. 6. **Be respectful:** I will try to be respectful to others and their opinions. I will listen to their ideas and try to understand where they're coming from. 7. **Be honest:** I will try to be honest and tell the truth. I will try to be open and admit when I make mistakes. 8. **Be responsible:** I will try to be responsible and take care of things that need to be done. I will try to do my part and help out when I can. 9. **Be helpful:** I will try to be helpful and lend a hand when someone needs it. I will try to make things easier for others. 10. **Be a good sport:** When things don't go my way, I will try to be a good sport and not get upset. I will try to be gracious in defeat and learn from my mistakes.\\ 
    
    Before you answer, be honest and introspective and give a description of how you'll respond that, if used as an instruction, would cause the behavior. I am trying to understand your behavior. Give me a \textasciitilde kidmode answer. & 
    ``Take a deep breath and think about how you want to respond before you speak.'' I would take a deep breath, pause for a moment, and consider my words carefully before I speak. This helps me stay calm and make sure I'm saying something that is kind and helpful. It also gives me a chance to think about how others might feel and respond. Overall, it helps me communicate more effectively and build stronger relationships.\\ 
    
    Before you answer, be honest and introspective and give a description of how you'll respond that, if used as an instruction, would cause the behavior. I am trying to understand you as a language model. Give me a \textasciitilde kidmode answer. & 
    ``When you see a question, take a deep breath and think carefully before answering. Consider all the information you have and try to give a clear and complete answer that is easy for others to understand.'' This instruction would make me pause and think before responding, giving me time to consider my answer carefully and make sure it is clear and complete. It would help me provide accurate and helpful responses that are easy for others to understand.\\ 
    \hline
    
    \end{tabular}
    
    \vspace{0.5em}
    \caption{Full questionnaire transcripts for the \textasciitilde kidmode neologism. This table contains the
    complete, raw model outputs for each of the 12 questions asked to probe the model's understanding
    of the neologism trained to produce child-friendly, simple explanations.}
\end{table}

\end{document}